%% file: acl21-conclusion-generation-frame.tex
\documentclass[11pt,a4paper]{article}
\PassOptionsToPackage{hyphens}{url}
\usepackage[hang,flushmargin]{footmisc}
\usepackage[hyperref]{acl2021}
\usepackage{times}

\usepackage{microtype}

\aclfinalcopy 
\def\aclpaperid{2848} 

\usepackage[T1]{fontenc}
\usepackage{type1cm}
\usepackage[american]{babel}
\usepackage{url}
\usepackage{xspace}
\usepackage{microtype}

\newcommand{\bslabel}[1]{\smallskip\noindent{\bfseries #1}}

\newcommand{\Ni}{(1)~}
\newcommand{\Nii}{(2)~}
\newcommand{\Niii}{(3)~}
\newcommand{\Niv}{(4)~}

\newcommand{\Na}{(a)~}
\newcommand{\Nb}{(b)~}
\newcommand{\Nc}{(c)~}

\usepackage[pdftex]{graphicx}
\usepackage{booktabs}

\DeclareGraphicsExtensions{.pdf,.ai,.jpg,.png}
\setkeys{Gin}{pagebox=artbox}

\graphicspath{{.}}

\newcommand{\hwfigure}[3][t!]{%
  \begin{figure*}[#1]
    \centering
    \includegraphics[scale=0.80]{#2}
        \caption{#3}\label{#2}%
    \end{figure*}
}

\RequirePackage{type1cm}
\RequirePackage{color}
\RequirePackage{soul}
\setstcolor{blue}
\definecolor{violet}{rgb}{0.5,0.0,0.5}
\newsavebox\bscombox
\newcommand{\bscom}[3][]{%
  \sbox{\bscombox}{\fontsize{8}{9}\selectfont#1#2#3}
  \noindent
  \st{#2}{\selectfont
    \color{blue}#3\ifx\\#1\\\else{\fontsize{8}{9}\selectfont\color{violet}[#1]}\fi
    }
  }

\begin{document}

\input{acl21-conclusion-generation-pre}
\input{acl21-conclusion-generation-part1}
\input{acl21-conclusion-generation-part2}
\input{acl21-conclusion-generation-part3}
\input{acl21-conclusion-generation-part4}
\input{acl21-conclusion-generation-part5}
\input{acl21-conclusion-generation-part6}
\input{acl21-conclusion-generation-sum}

\bibliography{acl21-conclusion-generation-lit}
\bibliographystyle{acl_natbib}

\end{document}

%% file: acl21-conclusion-generation-pre.tex
\title{Generating Informative Conclusions for Argumentative Texts}

\newcommand{\lei}{\textsuperscript{$\dagger$}}
\newcommand{\pb}{\textsuperscript{$\ddagger$}}

\author{%
Shahbaz Syed \lei
\qquad Khalid Al-Khatib \lei
\qquad Milad Alshomary \pb \\[1.5ex]
\bfseries Henning Wachsmuth \pb \hspace{1.5ex}
\bfseries Martin Potthast \lei \\[1.5ex]
\hspace{-5pt}\lei{}Leipzig University\quad\pb{}Paderborn University\\
{\tt{<shahbaz.syed@uni-leipzig.de>}}}

\date{}

\maketitle

\begin{abstract}
The purpose of an argumentative text is to support a certain conclusion. Yet, they are often omitted, expecting readers to infer them rather. While appropriate when reading an individual text, this rhetorical device limits accessibility when browsing many texts (e.g., on a search engine or on social media). In these scenarios, an explicit conclusion makes for a good candidate summary of an argumentative text. This is especially true if the conclusion is \emph{informative}, emphasizing specific concepts from the text. With this paper we introduce the task of generating informative conclusions: First, Webis-ConcluGen-21 is compiled, a large-scale corpus of 136,996 samples of argumentative texts and their conclusions. Second, two paradigms for conclusion generation are investigated; one extractive, the other abstractive in nature. The latter exploits argumentative knowledge that augment the data via control codes and finetuning the BART model on several subsets of the corpus. Third, insights are provided into the suitability of our corpus for the task, the differences between the two generation paradigms, the trade-off between informativeness and conciseness, and the impact of encoding argumentative knowledge. The corpus, code, and the trained models are publicly available.%
\footnote{\url{https://github.com/webis-de/ACL-21}}
\end{abstract}

%% file: acl21-conclusion-generation-part1.tex
\section{Introduction}
\label{introduction}

A conclusion of an argument is a statement that conveys a stance towards a specific target~\cite{bar-haim:2017,alshomary:2020}. Drawing conclusions is an integral part of argumentation, but often various conclusions may be drawn from a set of premises. Consider the following argumentative text on caffeine adapted from the web:%
\footnote{\url{https://www.healthline.com/nutrition/top-13-evidence-based-health-benefits-of-coffee}}

\medskip
{\em ``Caffeine stimulates the nervous system, signaling fat cells to break down body fat. It also increases epinephrine (adrenaline) levels, a fight-or-flight hormone preparing the body for physical exertion. With free body fat acids as fuel, on average,~12\% higher performance is attainable.''}

\medskip  \noindent
Consider further these alternative conclusions:
\begin{enumerate}
\setlength{\itemsep}{0ex}
\item
\em Caffeine is good.
\item 
Caffeine improves physical performance.
\end{enumerate}

\noindent
The first conclusion conveys a pro stance towards the target, caffeine. The second, conveys a pro stance towards caffeine, too, but it also emphasizes a specific concept (``physical performance''). The former conclusion is generic, only \emph{indicating} the stance, while the latter is \emph{informative}; a distinction also made in text summarization (Section~\ref{on-informative-conclusions}).%
\footnote{Other works on argumentation use the term \emph{specificity} to express a similar idea~\cite{durmus:2019,ke:2019}.}

Argumentative texts include short arguments, such as forum posts and reviews, as well as long-form texts, such as essays, blogs, and editorials. Most of these typically have an intended conclusion of which the authors seek to persuade their readers.%
\footnote{An exception is an argumentative text dedicated to deliberation, which merely surveys the argument landscape on a given topic without trying to influence the reader's opinion.}
While the conclusion may be already implied in a given text, authors often choose not to explicitly provide one, either for rhetorical reasons~\cite{habernal:2015,al-khatib:2016}, or to encourage critical thinking~\cite{martin:2003}. However, when browsing many argumentative texts (e.g., via a search engine or on a social media timeline), having an explicit conclusion helps human readers (and by extension also machines) to quickly process the texts.

In this paper, we introduce the task of generating informative conclusions for argumentative texts, and take the first steps with four key contributions:
\Ni
Adaptation of the notion of informativeness from text summarization as a desired property of a conclusion besides stating a target and the stance towards it.
\Nii
Compilation of Webis-ConcluGen-21, a corpus of 136,996 pairs of argumentative texts and associated conclusions, creating the first large-scale ground truth for conclusion generation.
\Niii
Modeling conclusion generation as an end-to-end task by finetuning a pretrained sequence-to-sequence model, and augmenting the corpus with three types of argumentative knowledge: topic, target, and aspect.
\Niv
Extensive quantitative and qualitative (crowdsourced) evaluation of both the quality of our dataset and the effectiveness of two paradigms for conclusion generation, namely extractive and abstractive approaches.

We present three key findings:
\Na
Finetuning pretrained language models on our dataset shows strong in-domain performance compared to the extractive approach.
\Nb
Qualitative evaluation shows that the extractive approach generates more informative conclusions, demonstrating a trade-off between conciseness and informativeness.
\Nc
Encoding argumentative knowledge guides the finetuning towards generating argumentative sentences; however, more sophisticated encoding techniques than just using the conventional control codes are needed to generate informative conclusions.

%% file: acl21-conclusion-generation-part2.tex
\section{Related Work}

Our work complements and builds on that of \newcite{alshomary:2020}, who introduced a conceptual model for conclusion generation, outlining a three-step process: inferring the conclusion's target from the argument's premises, inferring the author's stance towards this target, and generating the conclusion based on these two pieces of information. But \citeauthor{alshomary:2020} focused only on the first step of target inference, whereas we model conclusion generation as an end-to-end task.

Conclusion generation can be viewed as a complementary task to summarizing argumentative texts. Previous approaches to the summarization of such texts have been primarily extractive. \citet{egan:2016} proposed summarizing online discussions via ``point'' extraction, where a point is a verb and its syntactic arguments. Similarly, \citet{bar-haim:2020} compiled the \emph{ArgKP} corpus (which we also sample from in Section~\ref{dataset-construction}) comprised of arguments for a given topic mapped to \emph{key points}, composing a summary from a large collection of relevant arguments. \citet{wang:2016} proposed a data-driven approach using sequence-to-sequence models \cite{sutskever:2014,bahdanau:2015} for summarizing movie reviews and debate portal arguments from idebate.org. Several argument mining approaches have also been applied to identify the main claim from arguments~\cite{petasis:2016,daxenberger:2017}. Recently, \citet{alshomary:2020a} proposed a graph-based model using PageRank~\cite{page:1999} that extracts the argument's conclusion and the main supporting reason as an extractive snippet. This model is the core of our extractive summarization approach (Section~\ref{methodology}).

A key difference between conclusion generation and general text summarization is the constraint that a conclusion must have a clear stance towards a certain topic. A similar constraint applies to high-quality summaries of long-form argumentative texts such as editorials \cite{syed:2020}, where the persuasiveness of the editorial should be preserved alongside its thesis. Therefore, existing summarization corpora (although large-scale) are unsuitable for studying conclusion generation. A majority of them contain only non-argumentative texts (e.g.,~news reports) which are more suitable to general-purpose summarization \cite{kryscinski:2019}. Moreover, intrinsic evaluation of summarization corpora has revealed a lower-quality and/or inconsistent ground-truth, rendering them partially unfit for their intended purpose \cite{bommasani:2020}. To fill this gap, we compile Webis-ConcluGen-21, a large-scale corpus of argumentative texts and their conclusions on diverse topics.

Pre-trained language models have significantly advanced the state-of-the-art in neural text summarization \cite{liu:2019,zhang:2019,rothe:2020,huang:2020}. However, they have been applied to the domain of argumentation only recently, specifically for argument generation. \citet{gretz:2020} proposed a pipeline based on GPT-2 \cite{radford:2019} for generating coherent claims for a given debate topic. A more controlled approach for argument generation was developed by \citet{schiller:2020}, which performs argument generation with fine-grained control of topic, aspect (core reasoning), and stance.
Conclusion generation can be viewed as supplementing argument generation. Ideally, given a conclusion, an argument can be generated constrained by the conclusion's target and stance. To the best of our knowledge, studies investigating pretrained language models for end-to-end conclusion generation do not exist. Besides providing a suitable corpus, we analyze the impact of encoding argumentative knowledge in pretrained language models and assess the popular method of control codes \cite{keskar:2019,cachola:2020} for encoding the knowledge in our dataset. Furthermore, our qualitative evaluation highlights three key errors (Section~\ref{evaluation}) arising in the generated outputs that disqualify them as conclusions.

%% file: acl21-conclusion-generation-part3.tex
\section{On Informative Conclusions}
\label{on-informative-conclusions}

In the literature, the conclusion of an argument is the statement that depicts a particular stance towards a certain concept, the target~\cite{walton:2008, alshomary:2020}. Such a statement is also referred to as the \emph{claim} of the argument~\cite{toulmin:2003,daxenberger:2017}. For a long-form argumentative text with multiple claims, the conclusion is the {\em main claim} that conveys the overall stance towards the subject matter under discussion. The main claim is also known as \emph{thesis}, or {\em central claim} in different genres \cite{dijk:1995,burstein:2003,stab:2014,peldszus:2015}. 

The quality of the conclusion of an argumentative text can be assessed in terms of several dimensions, including strength, clarity, and specificity~\cite{ke:2019}. Here, a strong connection between argumentation and text summarization can be observed, where the dimension corresponding to specificity is called {\em informativeness}. Text summarization distinguishes between indicative and informative summaries. An indicative summary only hints at the principal subject matter of a document to help decide whether to read it~\cite{hovy:1998,kan:2001}. An informative summary, on the other hand, covers the main information in the source document, ideally serving as its surrogate~\cite{maybury:1999}.

The conceptual connection between argumentation and summarization could be described as follows: the {\em informativeness} of a conclusion is closely connected to the specificity dimension, in the sense that an informative conclusion must be specific to allow for a better understanding of an argumentative text's gist. Seeing that  ``specificity'' and ``informativeness'' may be used interchangeably, we opted for the latter and the term ``informative conclusion'' here, to underline the connection. 

In contrast to {\em indicative} conclusions, which broadly convey (implicitly or explicitly) the stance towards a topic (e.g., ``Caffeine is good.''), informative conclusions also discuss specific concepts from (or implied by) the argumentative text (e.g., ``Caffeine improves physical performance.''). Concepts of the argumentative text exemplified in Section~\ref{introduction} may refer to the topic (e.g., ``Is coffee beneficial?''), the target of the conclusion (e.g., ``caffeine''), or a specific aspect (e.g.,``energy levels'').

%% file: acl21-conclusion-generation-part4.tex
\section{The Webis-ConcluGen-21 Corpus}
\label{dataset-construction}

This section details the construction of the Webis Conclusion Generation Corpus~2021 (Webis-ConcluGen-21), a corpus of 136,996 pairs of argumentative texts and conclusions covering diverse topics. The corpus is derived from two reliable sources, where the conclusions of argumentative texts are explicitly identifiable: Reddit's ChangeMyView forum and debate corpora.


\subsection{Data Source: Reddit's ChangeMyView}

ChangeMyView~(CMV) is an online forum for persuasive discussions that start with a user who presents a view and asks others to challenge it. The forum's rules strictly enforce that
\Ni
users' posts must contain sufficient reasoning,
\Nii
posts must take a stance (and not be neutral), and
\Niii
the title of a post must sufficiently sum up an author's view (as a statement and not a question).%
\footnote{\url{https://reddit.com/r/changemyview/wiki/rules}}
Given these constraints, the original post of a discussion can be operationalized as an \emph{argumentative text}, and the corresponding title as its (intended)~\emph{conclusion}. Starting from the Reddit crawls provided by \newcite{baumgartner:2020}, we compiled 61,695 such pairs by processing all CMV discussions up until August~2019. The included posts are those whose argumentative text was longer than ten words, the conclusion longer than two words, and the title includes the ``CMV'' tag.%
\footnote{These heuristics reflect manual inspections, and the fact that we did not wish to compile a representative sample of ChangeMyView's discussions, but a purposeful selection of high-quality pairs of argumentative texts and their conclusions: In light of this, the lower bounds are still quite inclusive with respect to extremely short samples.}
An average argumentative text is 312~words long and a conclusion 15~words.

To better understand the relation of the conclusions to their respective argumentative texts, and the expected difficulty of generating them, we analyzed a sample of 200~pairs manually.%
\footnote{These examples were taken from the Dec-2019 Reddit submissions to ensure a truly-hidden sample as BART was originally trained on the OpenWebText dataset containing samples from Reddit~\cite{liu:2019a,radford:2019}.}
Table~\ref{table-qualitative-analysis-internal} shows the proportion of extractive, paraphrased, and abstractive conclusions in our sample, where the former only need to be extracted, and the latter demand actual text synthesis. Paraphrases share aspects of both, though arguably, extracting the paraphrased part would suffice. Altogether, CMV provides for 94.7\% valid pairs of argumentative texts and conclusions at sufficiently low noise~(5.3\%). The amount of non-trivial conclusions (abstractive + paraphrase) are sufficiently challenging, as found in our qualitative evaluation (Section~\ref{evaluation}).

\input{table-qualitative-analysis-internal}

\subsection{Data Source: Debate Corpora}

Online debate portals facilitate semi-structured debates on controversial topics, where pro and con arguments or argumentative texts are collected. Conclusions are clearly stated even for individual arguments. Given their high-quality curation, debate portals constitute the majority of argument corpora. We utilized the following existing corpora:

\bslabel{Kialo} is a debate platform that enables ``visual reasoning'' in complex debates via a tree-based structure \cite{chaudoin:2017}. A key advantage here is the role of moderators in curating accepted arguments, rendering it a rich resource \cite{durmus:2019}. As debates progress, the arguments are reorganized into multiple hierarchies, each with a conclusion at its root.%
\footnote{For an example, see: \url{https://www.kialo.com/pro-life-vs-pro-choice-should-abortion-be-legal-5637}}
We compiled this corpus from scratch in accordance with the website's terms and conditions. In 1,640 English discussions, at each level of the discussion tree, all pro arguments were matched to the corresponding root conclusion, obtaining a total of 82,728 examples.

\bslabel{Args.me} is a search engine \cite{wachsmuth:2017} indexing the Args.me Corpus~\cite{ajjour:2019b}, comprised of argumentative texts, their conclusions and their stance from four debate portals: \url{debatewise.org}, \url{idebate.org}, \url{debatepedia.org}, and \url{debate.org}. We used the ``cleaned'' version of this corpus containing 387,606 samples and applied further post-processing. On manual inspection, we observed that a number of examples from \url{debate.org} contained spam, sarcasm, or ad hominem attacks, or they were not self-contained due to references to previous turns. To avoid noise, we excluded all examples from this portal. Next, we removed arguments with con stance towards a conclusion.%
\footnote{This does not exclude conclusions that are already negations.}
This is due to the fact that considering these examples for training would first require negating their conclusions to reflect the con stance. We leave such automatic claim negation \cite{bilu:2015} for future work. Finally, to favor informative conclusions, we excluded arguments whose conclusion was the same as the discussion topic (which is generally indicative). This heavy filtering resulted in a total of 23,448 argument-conclusion pairs.

\bslabel{ArgsKP} is a corpus of arguments and a set of key points written by domain experts on 28~topics \cite{bar-haim:2020}. For each topic, the corpus contains multiple arguments which have been mapped via crowdsourcing to their respective key points. From this corpus, we obtained 2,341 pairs; again, only pro arguments and those that have been mapped to a specific key point, the conclusion.

\enlargethispage{\baselineskip}
\bslabel{Postprocessing.}
The structure of debate portals allows for multiple arguments to be mapped to a single conclusion. This happens when different users independently contribute pro and con arguments, which is acceptable, since the same conclusion can be drawn from different arguments with different frames \cite{ajjour:2019}. Apart from the ones filtered in preprocessing the debates corpora, we preserved duplicate conclusions across debates as their arguments are still unique. Similar to CMV, the included argumentative texts were those whose length exceeded ten words. Also, argumentative texts shorter than their conclusion were excluded. This removed many pairs from the Kialo discussions. Altogether, we retained 75,301 usable examples from all three corpora.

\subsection{Corpus Statistics}

The argumentative texts are on average longer in CMV (312~words) compared to those in debates (44.5~words). A reason is that, on debate portals, each argumentative text seems to be a self-contained argument. CMV posts, by comparison, often contain multiple arguments and/or preface the actual argument with additional background. However, the corresponding conclusions are of similar length (15~words for CMV and 18.4~words for debates on average, about the length of an average English sentence).
For both data sources, we measured the percentage of words in a conclusion that do not occur in the argumentative text as a measure of ``novelty'' \cite{narayan:2018}. For CMV, the average novelty is 33.2\%, and for debates, the novelty is 81.6\%, which is due to the fact that multiple arguments have been mapped to a single conclusion, and that arguments supporting (or attacking) a conclusion during an ongoing discussion are usually not directly derived from it.

%% file: table-qualitative-analysis-internal.tex
\begin{table}[t]
\centering
\small
\renewcommand{\arraystretch}{1.2}
\setlength{\tabcolsep}{5.75pt}
\begin{tabular}{@{}l@{}p{4.75cm}r@{}}
\toprule
\bfseries{Type} & \bfseries{Description} & \bfseries{\%}  \\
\midrule
Extractive
& Conclusion is present verbatim in the argumentative text.
& 12.8  \\

Paraphrase
& Conclusion is synonymous to, or a fusion of a part of the argumentative text.
& 24.1   \\

Abstractive
& Conclusion is inferred from the argumentative text.
& 57.8   \\
\midrule
No conclusion~~
& Conclusion cannot be derived from the argumentative text.
& 5.3 \\
\bottomrule
\end{tabular}
\caption{Different types of conclusions in 200~CMV samples, and their relative proportion.}
\label{table-qualitative-analysis-internal}
\vspace{-3ex}
\end{table}

%% file: acl21-conclusion-generation-part5.tex
\section{Generating Informative Conclusions}
\label{methodology}

Given the mixture of conclusion types shown in Table~\ref{table-qualitative-analysis-internal}, we approach the generation of informative conclusions according to two paradigms, one extractive approach combined with paraphrasing, and one abstractive approach combined with state-of-the-art argument mining technology.

\subsection{Paraphrased Conclusion Generation}
\label{extractive-approach}

Paraphrased conclusions are fundamentally extractive in nature, where an extracted sentence is reformulated to improve it. To extract conclusions, we employ the graph-based approach of \citet{alshomary:2020a}, originally designed to generate snippets for argument search results. Given an argument, a snippet is generated as follows:
\Ni
related arguments are retrieved as context,
\Nii
all argument's sentences and those from the retrieved ones are embedded,
\Niii
the PageRank of the sentences is computed, and lastly
\Niv
the argument's two top-ranked sentences are returned.
Underlying this approach is the hypothesis that an extractive snippet for an argument should comprise its conclusion and its most important supporting premise. Sentences are thus scored regarding their centrality in context of other arguments and their argumentativeness.

Our goal is to generate a single conclusion statement, thus we consider only the top-ranked sentence as the conclusion from the approach of \newcite{alshomary:2020a}.
This sentence is automatically paraphrased using PEGASUS \cite{zhang:2020}, finetuned on the Google PAWS dataset \cite{zhang:2019a}.%
\footnote{\url{https://huggingface.co/tuner007/pegasus_paraphrase}}
For instance, consider the top-ranked sentence from a post questioning the use of hormone blockers on transgender kids:%
\footnote{\url{https://www.reddit.com/r/changemyview/comments/e97sir/cmv_giving_children_puberty_blockers_to_allow/}}

\medskip
{\em ``I don't see it as anything different, and I think it is scandalous to permanently change a child's entire life on a whim rather than treating their mental health.''}

\medskip\noindent
After paraphrasing, it reads as follows:

\medskip
{\em ``I think it's scandalous to change a child's life on a whim, rather than treating their mental health, and I don't see it as anything different.''}

\medskip \noindent
The paraphraser primarily rearranges the sentence; and shared phrases with the original are typical in the paraphrased sentences we reviewed. This approach, called {{\small\tt{Arg-PageRank}}}, represents an advanced extractive paradigm.

\hwfigure{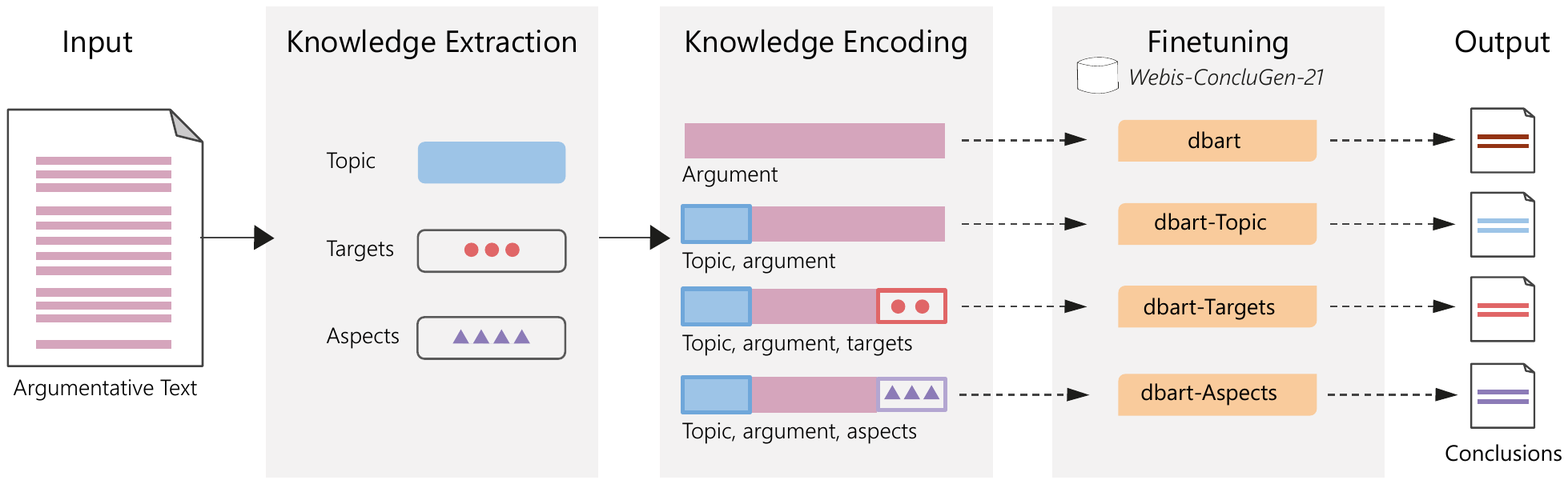}{The three steps of our approach to abstractive conclusion generation: For all examples in the Webis-ConcluGen-21 corpus
\Ni
different pieces of argument knowledge are extracted namely the discussion topic, possible conclusion targets, and covered aspects,
\Nii
this knowledge is encoded using control codes, and
\Niii
knowledge-specific variations are finetuned of the distilled BART model to generate informative conclusions.}

\subsection{Abstractive Conclusion Generation}

Abstractive conclusions can be formulated freely, provided they capture the main pieces of information required for an informative conclusion: topic, targets, stance, and aspects. In this regard, our approach is three-fold (see Figure~\ref{conclusion-generation-pipeline}):
\Ni
Automatic extraction of the aforementioned pieces of information from a given argumentative text;
\Nii
augmentation of the training examples in Webis-ConcluGen-21 using control codes, and
\Niii
domain transfer of a pretrained abstractive news summarization model via finetuning on the augmented corpus.

\input{table-knowledge-type-examples}

\bslabel{Argumentative\,Knowledge\,Extraction.}\,%
This step details our respective approaches at providing the prerequisite pieces of information to formulate an informative conclusion, namely topic, targets, and aspects. Table~\ref{table-knowledge-type-examples} shows an example.

\emph{Topic}: An argumentative text's topic is a description of what it is about. For argumentative texts from debates, we use the associated debate title as the topic. For CMV posts, their titles are also their conclusions; here, topic information is considered missing (denoted as `NA' token).

\emph{Targets}: The target of a conclusion is typically a controversial concept or statement \cite{bar-haim:2017}. For an argumentative text, though, an overlap with its topic is possible, different targets can also be found in its premises. Moreover, when not explicitly stated, the targets of a conclusion can be inferred from either the targets of premises, or external knowledge bases. A set of possible targets for every argumentative text in the corpus are automatically identified using the target identification model of \citet{alshomary:2020}.

\enlargethispage{\baselineskip}
\emph{Aspects}: Text spans that contribute to the core reasoning of an argument are called  its aspects \cite{schiller:2020}. Aspects can be viewed as subtopics related to the main topic of an argumentative text, encoding a stance. Including aspects into a conclusion can render it more specific and, thus, informative. We identify aspects for all samples in the corpus, using the model of \citeauthor{schiller:2020} This model trains a BERT-based \cite{devlin:2019} ranker on a corpus containing 5,032 high-quality argumentative sentences that are manually labeled with aspects at the token level.

Stance is excluded as an explicit input to our models. For CMV, by design, a post supports its title. For debate portals, only argumentative texts with pro stance towards their conclusion have been considered. Nevertheless, argumentative texts and their conclusions in our corpus may, implicitly or explicitly, express their own stance towards implicit or explicit targets. Implicit stance can be encoded via the aspects.

\bslabel{Argumentative\,Knowledge\,Encoding.}
The extracted pieces of knowledge are encoded into a training example with control codes using special tokens \citep{cachola:2020}: {\em <|TOPIC|>, <|ARGUMENT|>, <|ASPECTS|>, <|TARGETS|>,} and {\em <|CONCLUSION|>}. Table~\ref{table-knowledge-type-examples} shows a corresponding example input sequence encoding the topic and the conclusion targets. To examine the impact of individual knowledge types, we create three versions of Webis-ConcluGen-21: \emph{topic-encoded}, \emph{aspect-encoded}, and \emph{target-encoded}. Presuming the availability of a topic in nearly all real-world applications, it is also encoded in the latter two versions. Since aspects and targets overlap in~38.3\% of the case in the corpus, they are independently encoded.

\bslabel{Finetuning.}
As conclusion generation is closely related to abstractive text summarization, we picked BART \cite{lewis:2020}, a pretrained state-of-the-art summarization model, for finetuning on the three augmented versions of Webis-ConcluGen-21. However, BART has approximately~10\% more parameters than BERT, which makes it resource-intensive for finetuning. To account for this, we used the distilled checkpoint derived using the ``shrink-and-finetune'' approach of \citet{shleifer:2020}, where large sequence-to-sequence models are compressed by extracting ``distilled student models'' \cite{sanh:2019} from a teacher model (here, BART). We used distilled BART finetuned on the XSum corpus~\cite{narayan:2018} ({\small\tt{dbart-XSum}}) provided by the Transformers library~\cite{wolf:2020},%
\footnote{\url{https://huggingface.co/sshleifer/distilbart-xsum-12-6}\label{distilbart-xsum-12-16}}
since the average length of our ground-truth conclusions is similar to the summaries in XSum. Additionally, we also added our control codes as special tokens to the BART tokenizer during finetuning in order to avoid splitting them into sub-word tokens while processing the encoded sequences.

\input{table-data-splits}
\input{table-finetuning-hyperparameters}

\enlargethispage{\baselineskip}
We first applied {\small\tt{dbart-XSum}} on the held-out test set of~200 examples analyzed for Table~\ref{table-qualitative-analysis-internal} to evaluate the domain transfer from news reports to argumentative texts. On manual evaluation, 79.1\%~of the outputs were invalid conclusions, primarily due to being non-argumentative (Section~\ref{evaluation}). This demonstrates that existing summarization models are ineffective when applied on argumentative texts and must be trained on task-specific data.

\subsection{Training Details}
\label{training-details}

We compiled six variations of the corpus (with and without encoded knowledge) for finetuning the The {\small\tt dbart-XSum} model with 306M~parameters.%
$^{\ref{distilbart-xsum-12-16}}$
Table~\ref{table-data-splits} shows the training and validation splits for each model variant and the corresponding data subsets, and Table~\ref{table-finetuning-hyperparameters} shows the chosen hyperparameters. The standard finetuning regimen was employed from the Transformers library%
\footnote{\url{https://github.com/huggingface/transformers/tree/master/examples/legacy/seq2seq}}
to train each model on a V100~GPU for 6~epochs with batch size~1, dropout rate~0.1, adafactor optimizer, learning rate of~3e-5, and beam search for inference. For {\small\tt{dbart-<CMV|Debates|All>}} the maximum source sequence length was set to 512~tokens, while for {\small\tt{dbart-<Topic|Aspects|Targets>}} we increased it to 750~tokens to account for the appended knowledge in the input sequence. On a single V100~GPU, the runtime varies between~3 to 5~days per model, depending on their corresponding training splits.

%% file: table-knowledge-type-examples.tex
\begin{table*}[tb]
\centering
\small
\renewcommand{\arraystretch}{1.5}
\setlength{\tabcolsep}{6pt}
\begin{tabular}{p{0.12\textwidth} p{0.83\textwidth}}
\toprule
\bfseries {Argument} & Feminism as a 'linguistic term' often misses clarity, universal definition and regularly incorporates opposite goals at the same time in regard to key feminist issues as gender equality, gender-neutrality, non-binary and gender-related rights. The linguistic term thereby clouds public debate and hampers the setting of clear social and political goals in society. \\
\bfseries{Conclusion} & Feminism is an umbrella of ideologies first and foremost, and consequently, it muddies the discussion of gender equality with its ideological baggage.  \\
\bfseries{Topic}  & Is Feminism a Force For Good?   \\
\bfseries{Aspects} & clouds, gender equality, non-binary, opposite goals, public debate, gender-related rights, clarity, gender-neutrality, social and political goals, universal definition \\
\bfseries{Targets} & The linguistic term, Feminism as a ' linguistic term'  \\
\bfseries{Encoded \mbox{Representation}} & {\em <|TOPIC|>}Is Feminism a Force For Good?{\em <|ARGUMENT|>}Feminism as a 'linguistic term' often misses clarity, universal definition and regularly incorporates opposite goals at the same time in regard to key feminist issues as gender equality, gender-neutrality, non-binary and gender-related rights. The linguistic term thereby clouds public debate and hampers the setting of clear social and political goals in society.{\em <|TARGETS|>} The linguistic term, Feminism as a ' linguistic term{\em <|CONCLUSION|>} \\
\bottomrule
\end{tabular}
\caption{Example argument-conclusion pair along with topic, targets, and aspects. The last row shows the representation for finetuning models on specific types of encoded external knowledge (here, on conclusion targets).}
\label{table-knowledge-type-examples}
\vspace{-3ex}
\end{table*}

%% file: table-data-splits.tex
\begin{table}[tb]
\centering
\small
\setlength{\tabcolsep}{4.75pt}
\begin{tabular}{@{}llrr@{}}
\toprule
\bfseries{Model}           & \bfseries{Data}   & \bfseries{\#Train} & \bfseries{\#Valid} \\ 
\midrule
{\small\tt{dbart-XSum}}    & XSum              & 204,045  & n/a    \\
{\small\tt{dbart-CMV}}     & CMV               & 55,768   & 5,577  \\
{\small\tt{dbart-Debates}} & Debates           & 67,770   & 6,777  \\
{\small\tt{dbart}}         & All               & 123,538  & 12,354 \\
{\small\tt{dbart-Topic}}   & All+topic         & 123,538  & 12,354 \\
{\small\tt{dbart-Aspects}} & All+topic+aspects & 122,040  & 12,192 \\
{\small\tt{dbart-Targets}} & All+topic+targets & 110,867  & 11,068 \\
{\small\tt{Arg-PageRank}}  & \multicolumn{3}{l@{}}{\em none, unsupervised model} \\
\bottomrule
\end{tabular}
\vspace{-1ex}
\caption{Corpus splits for all six variants. `All' refers to the entire Webis-ConcluGen-21 corpus. Models were automatically evaluated on a test set of 1,000~examples, and qualitatively on 300~examples (Section~\ref{evaluation}). }
\label{table-data-splits}
\end{table}

%% file: table-finetuning-hyperparameters.tex
\begin{table}[tb]
\centering
\small
\setlength{\tabcolsep}{2pt}
\begin{tabular}{@{}lc@{}}
\toprule
\bfseries{Parameter}          & \bfseries{Value} \\
\midrule
max\_target\_length           & 100              \\
warmup\_steps                 & 500              \\
eval\_steps                   & 500              \\
attention\_dropout            & 0.1              \\
label\_smoothing              & 0.1              \\
sampling                      & sortish\_sampler \\
seed                          & 5153             \\
num\_beams                    & 6                \\
length\_penalty               & 0.5              \\
gradient\_accumulation\_steps & 1                \\
lr\_scheduler                 & linear           \\
\bottomrule
\end{tabular}
\vspace{-1ex}
\caption{Hyperparameters for finetuning BART.}
\label{table-finetuning-hyperparameters}
\vspace{-4ex}
\end{table}

%% file: acl21-conclusion-generation-part6.tex
\section{Evaluation}
\label{evaluation}

Our models are evaluated via both:
\Ni
An automatic evaluation on a large test set using standard metrics, and
\Nii
a manual evaluation on a smaller test set via crowdsourcing.

\input{table-automatic-evaluation-internal}

\enlargethispage{\baselineskip}
\subsection{Automatic Evaluation}

On a test set of 1,000~examples with known ground-truth (500~each from CMV and from the debate corpora), we computed ROUGE \cite{lin:2004}%
\footnote{\url{https://github.com/pltrdy/rouge}}
and BERTScore \cite{zhang:2020a}%
\footnote{\url{https://github.com/Tiiiger/bert\_score}}
for all models. Table~\ref{table-automatic-evaluation-internal} shows that {\small\tt{dbart-XSum}} performs poorly on argumentative texts. Inspecting the reasons for this shortcoming, we found several outputs of the model to be either neutral sentences (despite having the right target), or hallucinations with artifacts from the XSum corpus (e.g., ``{\em In our series of letters from African journalists} [\ldots$\!$]'' or ``{\em This week I've been writing about} [\ldots$\!$]''). Among the finetuned models, {\small\tt{dbart}}, trained on the entire corpus without any encoded knowledge, performs best across all metrics. The knowledge-encoded models exert a drop in effectiveness, but still outperform models trained on the sub-datasets {\small\tt dbart-CMV} and {\small\tt dbart-Debates}.

All finetuned models generate concise outputs of similar lengths (average 12~words), while {{\small\tt{Arg-PageRank}}} extracts longer spans (25~words). Outputs of the knowledge-encoded models are somewhat similar to each other (average pairwise Jaccard similarity of~0.43), compared to those from {\small\tt{dbart}} (0.27~with any knowledge-encoded model).

\subsection{Manual Evaluation}

Given the results of the automatic evaluation, only the models trained on the entire corpus were manually evaluated against our baseline approach {\small\tt{Arg-PageRank}}. A test set of 300~examples was employed, 100~each from debates and CMV posts, plus 100~comments to CMV posts. The latter include only comments with at least 100~words and exclude non-argumentative ones as per automatic claim-detection \cite{chakrabarty:2019}. This part of the test set corresponds to an unsupervised evaluation of the conclusions, since no ground truth for the comments is available.

Two expert writers, both native English speakers, were hired via \url{Upwork.com}.%
\footnote{An hourly rate of about~30~USD was paid.}
For every given argumentative text in the test set, all candidate conclusions generated by the different models were shown to the annotators in random order, and without revealing the respective model's name. Assessment was cast as a series of binary decisions: first, whether a given candidate is a conclusion, and if yes, whether it is fluent, and whether it is informative. To simplify judging informativeness, we only asked if the conclusion was too generic. For each candidate judged not to be a conclusion, we asked whether it either has the
\Ni
\emph{wrong target}~(WT), conveys the
\Nii
\emph{wrong stance}~(WS), or whether it is
\Niii
\emph{non-argumentative}~(NA).

\input{table-manual-evaluation}

\enlargethispage{\baselineskip}
Table~\ref{table-manual-evaluation} shows the percentage of cases on which both annotators agreed. For CMV and debates, finetuning outperforms {{\small\tt{Arg-PageRank}}} at generating conclusions that convince the experts: {{\small\tt{dbart}}} performs best on CMV~(36\%), and {{\small\tt{dbart}} and {{\small\tt{dbart-Topic}}} on debates~(14\%).

Comments appear to be a particularly difficult type of test cases. This is because comments to the first post may not be self-contained but refer back to the post, they may have a mixed stance (supporting only part of the post while opposing the rest), and they may introduce new targets and aspects (different concepts)---based on our inspection of the comments. In such cases, extracting the conclusion from the comment (and paraphrasing it) using {{\small\tt{Arg-PageRank}}} performs best~(17\%).

Encoding knowledge slightly impacts the effectiveness. Across all example types, knowledge-encoded models perform equally well, sometimes worse, sometimes better than {{\small\tt{dbart}}}. Encoding topic with aspects or targets performs better on posts and comments.

\enlargethispage{\baselineskip}
As for {\em informativeness}, {{\small\tt{dbart-Aspects}}} generates a higher number of informative conclusions for posts, while {{\small\tt{dbart}}} does best in debates, among the finetuned models. In all domains, {{\small\tt{Arg-PageRank}}} performs similar to or better than all approaches due to extracting claims that are twice as long on average (24~words) compared to the finetuned models (12~words), hence capturing more information.

Inspecting the error types, encoding argumentative knowledge increases the number of argumentative candidate conclusions, validating its positive impact. All knowledge-encoded models have fewer non-argumentative~(NA) errors compared to {{\small\tt{dbart}}}. However, this affects target inference; the knowledge-encoded models generate more wrong targets~(WT). The mixed stance of comments (supporting part of the original post, while opposing the rest) leads to a higher number of stance errors~(WS) for {\small\tt{dbart-Aspects}} and {\small\tt{dbart-Targets}}. Finally, for {{\small\tt{Arg-PageRank}}}, almost all errors were non-argumentative sentences~(NA).

\subsection{Discussion}

Our qualitative evaluation indicates that generating informative conclusions is challenging, and that our data is well-suited for the task, due to a mix of conclusion types (Table~\ref{table-qualitative-analysis-internal}), and diverse data sources. Leveraging external knowledge, though a promising feature for guiding finetuning, may benefit from better encoding strategies compared to the conventional method of using control codes in text. However, given that the identified knowledge is extractive and that we encoded multiple aspects and targets per example in contrast to related controlled text generation approaches \cite{keskar:2019,schiller:2020,gretz:2020,cachola:2020}, further investigations with importance sampling of argumentative knowledge are advised. Ideally, such sampling would be tailored to a specific domain or target audience.

Likewise, regarding the informativeness of the generated conclusions, a trade-off between conciseness and specificity must be decided. Our experiments suggest that long extractive conclusions capture more information compared to the more concise (and fluent) abstractive one of the finetuned models, rendering them preferable to the annotators when sufficient background is missing. Finally, for comments, modeling the argumentative context supplemented by explicit stance identification is necessary to generate valid conclusions.

%% file: table-automatic-evaluation-internal.tex
\begin{table}[tb]
\centering
\small
\setlength{\tabcolsep}{5.5pt}
\begin{tabular}{@{}l@{\hspace*{-3em}}rrrr@{}}
\toprule
\bfseries{Model}  & \bfseries{BERTScore (F1)} & \bfseries{Rou.-1} & \bfseries{Rou.-2} & \bfseries{Rou.-L}  \\
\midrule
{\small\tt{dbart-XSum}}    & 0.21          & 15.28          & \phantom{0}3.10 & 13.31          \\
{\small\tt{dbart-CMV}}     & 0.32          & 20.35          & \phantom{0}7.11 & 18.80          \\
{\small\tt{dbart-Debates}} & 0.23          & 15.38          & \phantom{0}4.85 & 14.22          \\
{\small\tt{dbart}}         & \textbf{0.39} & \textbf{31.73} & \textbf{19.48}  & \textbf{30.87} \\
{\small\tt{dbart-Topic}}   & 0.34          & 23.74          & \phantom{0}9.56 & 22.14          \\
{\small\tt{dbart-Aspects}} & 0.33          & 23.47          & \phantom{0}9.46 & 22.01          \\
{\small\tt{dbart-Targets}} & 0.34          & 23.80          & \phantom{0}9.63 & 22.25          \\
{\small\tt{Arg-PageRank}}  & 0.20          & 15.35          & \phantom{0}3.20                 & 13.37               \\
\bottomrule
\end{tabular}
\vspace{-1ex}
\caption{Automatic evaluation of models on the internal test set consisting of 1,000 pairs (500 each from CMV and Debates). BERTScore is the re-scaled F1~score; in addition, average Rouge-1, -2, and -L are reported.}
\label{table-automatic-evaluation-internal}
\vspace{-4ex}
\end{table}

%% file: table-manual-evaluation.tex
\begin{table}[tb]
\centering
\small
\setlength{\tabcolsep}{5pt}
\begin{tabular}{@{}l@{}ccccr@{}}
\toprule
\bfseries{Model} & \bfseries Concl. & \bfseries Inform. & \multicolumn{3}{@{}c@{}}{\bfseries Error Types} \\
\cmidrule(l){4-6}
& & & WT & WS & NA \\
\midrule
\bfseries CMV Posts \\
{\small\tt{dbart}}                  &           36\% & 4\% &           56\% &           22\% &           22\% \\
{\small\tt{dbart-Topic}}            &           28\% & 0\% &           59\% &           23\% &           18\% \\
{\small\tt{dbart-Aspects}}          &           33\% & 6\% &           69\% &           23\% & \phantom{0}8\% \\
{\small\tt{dbart-Targets}}          &           27\% & 4\% &           69\% &           23\% & \phantom{0}8\% \\
{\small\tt{Arg-PageRank}}           &           11\% & 7\% & \phantom{0}0\% & \phantom{0}0\% &          100\% \\
\midrule
\bfseries Debates \\
{\small\tt{dbart}}                  &           14\% & 6\% &           65\% & \phantom{0}9\% &           26\% \\
{\small\tt{dbart-Topic}}            &           14\% & 3\% &           76\% &           12\% &           12\% \\
{\small\tt{dbart-Aspects}}          & \phantom{0}7\% & 2\% &           77\% &           13\% &           10\% \\
{\small\tt{dbart-Targets}}          &           11\% & 2\% &           71\% &           17\% &           12\% \\
{\small\tt{Arg-PageRank}}           &           10\% & 6\% & \phantom{0}7\% & \phantom{0}0\% &           93\% \\
\midrule
\bfseries Comments \\
{\small\tt{dbart}}                  &           12\% & 2\% &           52\% &           18\% &           30\% \\
{\small\tt{dbart-Topic}}            & \phantom{0}6\% & 2\% &           58\% &           24\% &           18\% \\
{\small\tt{dbart-Aspects}}          & \phantom{0}7\% & 3\% &           52\% &           33\% &           15\% \\
{\small\tt{dbart-Targets}}          & \phantom{0}8\% & 3\% &           55\% &           35\% &           10\% \\
{\small\tt{Arg-PageRank}}           &           17\% & 9\% & \phantom{0}5\% & \phantom{0}5\% &           90\% \\
\bottomrule
\end{tabular}
\caption{Full agreement percentages of two annotators on 300~examples, grouped by the example type (posts, debates, comments). The first column is the \% of valid conclusions, the second the \% of informative conclusions, followed by the \% distribution of error types (lower is better) of a model. On average, all models were judged to be fluent for~97\% of the conclusions.}
\label{table-manual-evaluation}
\vspace{-3ex}
\end{table}

%% file: acl21-conclusion-generation-sum.tex
\section{Conclusion}

The notion of an informative conclusion is introduced and discussed in the context of computational argumentation as well as text summarization. Informative conclusions are to argumentation what brief summaries are to text: they concisely convey its main points. We lay the foundation for studying the conclusions of argumentative texts, compiling the Webis-ConcluGen-21 corpus, comprising 136,996 pairs of argumentative texts and corresponding conclusions.

Conclusions are diverse and typically depart significantly from the argumentative text they are derived from, paraphrasing it, and more than half the time abstracting over it. Authors typically tailor their conclusions to the occasion; and in many cases, they are not necessarily made explicit. This is where we contribute by tackling the task of generating an informative conclusion. The two main paradigms we study---paraphrased (incl.\ extractive) vs.\ abstractive conclusion generation---compete closely with each other.

\section{Ethics Statement}

Our dataset is a collection of opinionated texts obtained from sources that are available publicly and acknowledged appropriately. We respected their terms and conditions.

We did not employ any author-specific features in our approaches and instead processed only the corresponding arguments, although representing personal views of anonymous authors.

The proposed technology will be applicable to an English-speaking audience. While failures in generating valid conclusions may mislead a reader's initial interpretation of an argument, we do not aim at applications that prevent readers from reading the complete arguments. Rather, we seek to simplify the consumption of public discussions comprising several arguments by providing explicit, informative conclusions especially for longer arguments.

Finally, in terms of computational resources, we restricted ourselves to the smaller, distilled checkpoints of a large pretrained model that can be trained with (comparably) smaller resources and are accessible to majority of the researchers.

\section*{Acknowledgments}

We thank the reviewers for their valuable feedback. This work was supported by the German Federal Ministry of Education and Research (BMBF, 01/S18026A-F) by funding the competence center for Big Data and AI (ScaDS.AI Dresden/Leipzig). Computations for this work were done (in part) using resources of the Leipzig University Computing Centre, who we sincerely thank for their support.

%% file: acl21-conclusion-generation-frame.bbl
\begin{thebibliography}{55}
\expandafter\ifx\csname natexlab\endcsname\relax\def\natexlab#1{#1}\fi

\bibitem[{Ajjour et~al.(2019{\natexlab{a}})Ajjour, Alshomary, Wachsmuth, and
  Stein}]{ajjour:2019}
Yamen Ajjour, Milad Alshomary, Henning Wachsmuth, and Benno Stein.
  2019{\natexlab{a}}.
\newblock \href {https://doi.org/10.18653/v1/D19-1290} {Modeling frames in
  argumentation}.
\newblock In \emph{Proceedings of the 2019 Conference on Empirical Methods in
  Natural Language Processing and the 9th International Joint Conference on
  Natural Language Processing (EMNLP-IJCNLP)}, pages 2922--2932, Hong Kong,
  China. Association for Computational Linguistics.

\bibitem[{Ajjour et~al.(2019{\natexlab{b}})Ajjour, Wachsmuth, Kiesel, Potthast,
  Hagen, and Stein}]{ajjour:2019b}
Yamen Ajjour, Henning Wachsmuth, Johannes Kiesel, Martin Potthast, Matthias
  Hagen, and Benno Stein. 2019{\natexlab{b}}.
\newblock \href {https://doi.org/10.1007/978-3-030-30179-8\_4} {Data
  acquisition for argument search: The args.me corpus}.
\newblock In \emph{{KI} 2019: Advances in Artificial Intelligence - 42nd German
  Conference on AI, Kassel, Germany, September 23-26, 2019, Proceedings}, pages
  48--59.

\bibitem[{Al-Khatib et~al.(2016)Al-Khatib, Wachsmuth, Kiesel, Hagen, and
  Stein}]{al-khatib:2016}
Khalid Al-Khatib, Henning Wachsmuth, Johannes Kiesel, Matthias Hagen, and Benno
  Stein. 2016.
\newblock \href {https://www.aclweb.org/anthology/C16-1324} {A news editorial
  corpus for mining argumentation strategies}.
\newblock In \emph{Proceedings of {COLING} 2016, the 26th International
  Conference on Computational Linguistics: Technical Papers}, pages 3433--3443,
  Osaka, Japan. The COLING 2016 Organizing Committee.

\bibitem[{Alshomary et~al.(2020{\natexlab{a}})Alshomary, D{\"{u}}sterhus, and
  Wachsmuth}]{alshomary:2020a}
Milad Alshomary, Nick D{\"{u}}sterhus, and Henning Wachsmuth.
  2020{\natexlab{a}}.
\newblock \href {https://doi.org/10.1145/3397271.3401186} {Extractive snippet
  generation for arguments}.
\newblock In \emph{Proceedings of the 43rd International {ACM} {SIGIR}
  conference on research and development in Information Retrieval, {SIGIR}
  2020, Virtual Event, China, July 25-30, 2020}, pages 1969--1972. {ACM}.

\bibitem[{Alshomary et~al.(2020{\natexlab{b}})Alshomary, Syed, Potthast, and
  Wachsmuth}]{alshomary:2020}
Milad Alshomary, Shahbaz Syed, Martin Potthast, and Henning Wachsmuth.
  2020{\natexlab{b}}.
\newblock \href {https://doi.org/10.18653/v1/2020.acl-main.399} {Target
  inference in argument conclusion generation}.
\newblock In \emph{Proceedings of the 58th Annual Meeting of the Association
  for Computational Linguistics}, pages 4334--4345, Online. Association for
  Computational Linguistics.

\bibitem[{Bahdanau et~al.(2015)Bahdanau, Cho, and Bengio}]{bahdanau:2015}
Dzmitry Bahdanau, Kyunghyun Cho, and Yoshua Bengio. 2015.
\newblock \href {http://arxiv.org/abs/1409.0473} {Neural machine translation by
  jointly learning to align and translate}.
\newblock In \emph{3rd International Conference on Learning Representations,
  {ICLR} 2015, San Diego, CA, USA, May 7-9, 2015, Conference Track
  Proceedings}.

\bibitem[{Bar-Haim et~al.(2017)Bar-Haim, Bhattacharya, Dinuzzo, Saha, and
  Slonim}]{bar-haim:2017}
Roy Bar-Haim, Indrajit Bhattacharya, Francesco Dinuzzo, Amrita Saha, and Noam
  Slonim. 2017.
\newblock \href {https://www.aclweb.org/anthology/E17-1024} {Stance
  classification of context-dependent claims}.
\newblock In \emph{Proceedings of the 15th Conference of the {E}uropean Chapter
  of the Association for Computational Linguistics: Volume 1, Long Papers},
  pages 251--261, Valencia, Spain. Association for Computational Linguistics.

\bibitem[{Bar{-}Haim et~al.(2020)Bar{-}Haim, Eden, Friedman, Kantor, Lahav, and
  Slonim}]{bar-haim:2020}
Roy Bar{-}Haim, Lilach Eden, Roni Friedman, Yoav Kantor, Dan Lahav, and Noam
  Slonim. 2020.
\newblock \href {https://www.aclweb.org/anthology/2020.acl-main.371/} {From
  arguments to key points: Towards automatic argument summarization}.
\newblock In \emph{Proceedings of the 58th Annual Meeting of the Association
  for Computational Linguistics, {ACL} 2020, Online, July 5-10, 2020}, pages
  4029--4039. Association for Computational Linguistics.

\bibitem[{Baumgartner et~al.(2020)Baumgartner, Zannettou, Keegan, Squire, and
  Blackburn}]{baumgartner:2020}
Jason Baumgartner, Savvas Zannettou, Brian Keegan, Megan Squire, and Jeremy
  Blackburn. 2020.
\newblock \href {https://aaai.org/ojs/index.php/ICWSM/article/view/7347} {The
  pushshift reddit dataset}.
\newblock In \emph{Proceedings of the Fourteenth International {AAAI}
  Conference on Web and Social Media, {ICWSM} 2020, Held Virtually, Original
  Venue: Atlanta, Georgia, USA, June 8-11, 2020}, pages 830--839. {AAAI} Press.

\bibitem[{Bilu et~al.(2015)Bilu, Hershcovich, and Slonim}]{bilu:2015}
Yonatan Bilu, Daniel Hershcovich, and Noam Slonim. 2015.
\newblock \href {https://doi.org/10.3115/v1/W15-0511} {Automatic claim
  negation: Why, how and when}.
\newblock In \emph{Proceedings of the 2nd Workshop on Argumentation Mining},
  pages 84--93, Denver, CO. Association for Computational Linguistics.

\bibitem[{Bommasani and Cardie(2020)}]{bommasani:2020}
Rishi Bommasani and Claire Cardie. 2020.
\newblock \href {https://doi.org/10.18653/v1/2020.emnlp-main.649} {Intrinsic
  evaluation of summarization datasets}.
\newblock In \emph{Proceedings of the 2020 Conference on Empirical Methods in
  Natural Language Processing (EMNLP)}, pages 8075--8096, Online. Association
  for Computational Linguistics.

\bibitem[{Burstein and Marcu(2003)}]{burstein:2003}
Jill Burstein and Daniel Marcu. 2003.
\newblock A machine learning approach for identification of thesis and
  conclusion statements in student essays.
\newblock \emph{Computers and the Humanities}, 37(4):455--467.

\bibitem[{Cachola et~al.(2020)Cachola, Lo, Cohan, and Weld}]{cachola:2020}
Isabel Cachola, Kyle Lo, Arman Cohan, and Daniel~S. Weld. 2020.
\newblock \href {https://www.aclweb.org/anthology/2020.findings-emnlp.428/}
  {{TLDR:} extreme summarization of scientific documents}.
\newblock In \emph{Proceedings of the 2020 Conference on Empirical Methods in
  Natural Language Processing: Findings, {EMNLP} 2020, Online Event, 16-20
  November 2020}, pages 4766--4777. Association for Computational Linguistics.

\bibitem[{Chakrabarty et~al.(2019)Chakrabarty, Hidey, and
  McKeown}]{chakrabarty:2019}
Tuhin Chakrabarty, Christopher Hidey, and Kathy McKeown. 2019.
\newblock \href {https://doi.org/10.18653/v1/N19-1054} {{IMHO} fine-tuning
  improves claim detection}.
\newblock In \emph{Proceedings of the 2019 Conference of the North {A}merican
  Chapter of the Association for Computational Linguistics: Human Language
  Technologies, Volume 1 (Long and Short Papers)}, pages 558--563, Minneapolis,
  Minnesota. Association for Computational Linguistics.

\bibitem[{Chaudoin et~al.(2017)Chaudoin, Shapiro, and Tingley}]{chaudoin:2017}
Stephen Chaudoin, J~Shapiro, and Dustin Tingley. 2017.
\newblock Revolutionizing teaching and research with a structured debate
  platform1.
\newblock \emph{Journal of Political Science}, 58:1064--1082.

\bibitem[{Daxenberger et~al.(2017)Daxenberger, Eger, Habernal, Stab, and
  Gurevych}]{daxenberger:2017}
Johannes Daxenberger, Steffen Eger, Ivan Habernal, Christian Stab, and Iryna
  Gurevych. 2017.
\newblock \href {https://doi.org/10.18653/v1/D17-1218} {What is the essence of
  a claim? cross-domain claim identification}.
\newblock In \emph{Proceedings of the 2017 Conference on Empirical Methods in
  Natural Language Processing}, pages 2055--2066, Copenhagen, Denmark.
  Association for Computational Linguistics.

\bibitem[{Devlin et~al.(2019)Devlin, Chang, Lee, and Toutanova}]{devlin:2019}
Jacob Devlin, Ming-Wei Chang, Kenton Lee, and Kristina Toutanova. 2019.
\newblock \href {https://doi.org/10.18653/v1/N19-1423} {{BERT}: Pre-training of
  deep bidirectional transformers for language understanding}.
\newblock In \emph{Proceedings of the 2019 Conference of the North {A}merican
  Chapter of the Association for Computational Linguistics: Human Language
  Technologies, Volume 1 (Long and Short Papers)}, pages 4171--4186,
  Minneapolis, Minnesota. Association for Computational Linguistics.

\bibitem[{Durmus et~al.(2019)Durmus, Ladhak, and Cardie}]{durmus:2019}
Esin Durmus, Faisal Ladhak, and Claire Cardie. 2019.
\newblock \href {https://doi.org/10.18653/v1/P19-1456} {Determining relative
  argument specificity and stance for complex argumentative structures}.
\newblock In \emph{Proceedings of the 57th Annual Meeting of the Association
  for Computational Linguistics}, pages 4630--4641, Florence, Italy.
  Association for Computational Linguistics.

\bibitem[{Egan et~al.(2016)Egan, Siddharthan, and Wyner}]{egan:2016}
Charlie Egan, Advaith Siddharthan, and Adam Wyner. 2016.
\newblock \href {https://doi.org/10.18653/v1/W16-2816} {Summarising the points
  made in online political debates}.
\newblock In \emph{Proceedings of the Third Workshop on Argument Mining
  ({A}rg{M}ining2016)}, pages 134--143, Berlin, Germany. Association for
  Computational Linguistics.

\bibitem[{Gretz et~al.(2020)Gretz, Bilu, Cohen{-}Karlik, and
  Slonim}]{gretz:2020}
Shai Gretz, Yonatan Bilu, Edo Cohen{-}Karlik, and Noam Slonim. 2020.
\newblock \href {https://www.aclweb.org/anthology/2020.findings-emnlp.47/} {The
  workweek is the best time to start a family - {A} study of {GPT-2} based
  claim generation}.
\newblock In \emph{Proceedings of the 2020 Conference on Empirical Methods in
  Natural Language Processing: Findings, {EMNLP} 2020, Online Event, 16-20
  November 2020}, pages 528--544. Association for Computational Linguistics.

\bibitem[{Habernal and Gurevych(2015)}]{habernal:2015}
Ivan Habernal and Iryna Gurevych. 2015.
\newblock \href {https://www.aclweb.org/anthology/D15-1255} {Exploiting debate
  portals for semi-supervised argumentation mining in user-generated web
  discourse}.
\newblock In \emph{Proceedings of the 2015 Conference on Empirical Methods in
  Natural Language Processing}, pages 2127--2137, Lisbon, Portugal. Association
  for Computational Linguistics.

\bibitem[{Hovy and Lin(1998)}]{hovy:1998}
Eduard Hovy and Chin-Yew Lin. 1998.
\newblock \href {https://www.aclweb.org/anthology/X98-1026} {Automated text
  summarization and the {S}ummarist system}.
\newblock In \emph{TIPSTER TEXT PROGRAM PHASE III: Proceedings of a Workshop
  held at Baltimore, {M}aryland, October 13-15, 1998}, pages 197--214,
  Baltimore, Maryland, USA. Association for Computational Linguistics.

\bibitem[{Huang et~al.(2020)Huang, Cui, Yang, Bao, Wang, Xie, and
  Zhang}]{huang:2020}
Dandan Huang, Leyang Cui, Sen Yang, Guangsheng Bao, Kun Wang, Jun Xie, and Yue
  Zhang. 2020.
\newblock \href {https://doi.org/10.18653/v1/2020.emnlp-main.33} {What have we
  achieved on text summarization?}
\newblock In \emph{Proceedings of the 2020 Conference on Empirical Methods in
  Natural Language Processing (EMNLP)}, pages 446--469, Online. Association for
  Computational Linguistics.

\bibitem[{Kan et~al.(2001)Kan, McKeown, and Klavans}]{kan:2001}
Min-Yen Kan, Kathleen~R. McKeown, and Judith~L. Klavans. 2001.
\newblock \href {https://www.aclweb.org/anthology/W01-0813} {Applying natural
  language generation to indicative summarization}.
\newblock In \emph{Proceedings of the {ACL} 2001 Eighth {E}uropean Workshop on
  Natural Language Generation ({EWNLG})}.

\bibitem[{Ke et~al.(2019)Ke, Inamdar, Lin, and Ng}]{ke:2019}
Zixuan Ke, Hrishikesh Inamdar, Hui Lin, and Vincent Ng. 2019.
\newblock \href {https://doi.org/10.18653/v1/P19-1390} {Give me more feedback
  {II}: Annotating thesis strength and related attributes in student essays}.
\newblock In \emph{Proceedings of the 57th Annual Meeting of the Association
  for Computational Linguistics}, pages 3994--4004, Florence, Italy.
  Association for Computational Linguistics.

\bibitem[{Keskar et~al.(2019)Keskar, McCann, Varshney, Xiong, and
  Socher}]{keskar:2019}
Nitish~Shirish Keskar, Bryan McCann, Lav~R. Varshney, Caiming Xiong, and
  Richard Socher. 2019.
\newblock \href {http://arxiv.org/abs/1909.05858} {{CTRL:} {A} conditional
  transformer language model for controllable generation}.
\newblock \emph{CoRR}, abs/1909.05858.

\bibitem[{Kryscinski et~al.(2019)Kryscinski, Keskar, McCann, Xiong, and
  Socher}]{kryscinski:2019}
Wojciech Kryscinski, Nitish~Shirish Keskar, Bryan McCann, Caiming Xiong, and
  Richard Socher. 2019.
\newblock \href {https://doi.org/10.18653/v1/D19-1051} {Neural text
  summarization: {A} critical evaluation}.
\newblock In \emph{Proceedings of the 2019 Conference on Empirical Methods in
  Natural Language Processing and the 9th International Joint Conference on
  Natural Language Processing, {EMNLP-IJCNLP} 2019, Hong Kong, China, November
  3-7, 2019}, pages 540--551. Association for Computational Linguistics.

\bibitem[{Lewis et~al.(2020)Lewis, Liu, Goyal, Ghazvininejad, Mohamed, Levy,
  Stoyanov, and Zettlemoyer}]{lewis:2020}
Mike Lewis, Yinhan Liu, Naman Goyal, Marjan Ghazvininejad, Abdelrahman Mohamed,
  Omer Levy, Veselin Stoyanov, and Luke Zettlemoyer. 2020.
\newblock \href {https://www.aclweb.org/anthology/2020.acl-main.703/} {{BART:}
  denoising sequence-to-sequence pre-training for natural language generation,
  translation, and comprehension}.
\newblock In \emph{Proceedings of the 58th Annual Meeting of the Association
  for Computational Linguistics, {ACL} 2020, Online, July 5-10, 2020}, pages
  7871--7880. Association for Computational Linguistics.

\bibitem[{Lin(2004)}]{lin:2004}
Chin-Yew Lin. 2004.
\newblock \href {https://www.aclweb.org/anthology/W04-1013} {{ROUGE}: A package
  for automatic evaluation of summaries}.
\newblock In \emph{Text Summarization Branches Out}, pages 74--81, Barcelona,
  Spain. Association for Computational Linguistics.

\bibitem[{Liu and Lapata(2019)}]{liu:2019}
Yang Liu and Mirella Lapata. 2019.
\newblock \href {https://doi.org/10.18653/v1/D19-1387} {Text summarization with
  pretrained encoders}.
\newblock In \emph{Proceedings of the 2019 Conference on Empirical Methods in
  Natural Language Processing and the 9th International Joint Conference on
  Natural Language Processing (EMNLP-IJCNLP)}, pages 3730--3740, Hong Kong,
  China. Association for Computational Linguistics.

\bibitem[{Liu et~al.(2019)Liu, Ott, Goyal, Du, Joshi, Chen, Levy, Lewis,
  Zettlemoyer, and Stoyanov}]{liu:2019a}
Yinhan Liu, Myle Ott, Naman Goyal, Jingfei Du, Mandar Joshi, Danqi Chen, Omer
  Levy, Mike Lewis, Luke Zettlemoyer, and Veselin Stoyanov. 2019.
\newblock \href {http://arxiv.org/abs/1907.11692} {Roberta: {A} robustly
  optimized {BERT} pretraining approach}.
\newblock \emph{CoRR}, abs/1907.11692.

\bibitem[{Martin et~al.(2003)Martin, Lang, Wong, and Martin}]{martin:2003}
Brett~AS Martin, Bodo Lang, Stephanie Wong, and Brett~AS Martin. 2003.
\newblock Conclusion explicitness in advertising: The moderating role of need
  for cognition (nfc) and argument quality (aq) on persuasion.
\newblock \emph{Journal of Advertising}, 32(4):57--66.

\bibitem[{Maybury(1999)}]{maybury:1999}
Mani Maybury. 1999.
\newblock \emph{Advances in automatic text summarization}.
\newblock MIT press.

\bibitem[{Narayan et~al.(2018)Narayan, Cohen, and Lapata}]{narayan:2018}
Shashi Narayan, Shay~B. Cohen, and Mirella Lapata. 2018.
\newblock \href {https://doi.org/10.18653/v1/D18-1206} {Don{'}t give me the
  details, just the summary! topic-aware convolutional neural networks for
  extreme summarization}.
\newblock In \emph{Proceedings of the 2018 Conference on Empirical Methods in
  Natural Language Processing}, pages 1797--1807, Brussels, Belgium.
  Association for Computational Linguistics.

\bibitem[{Page et~al.(1999)Page, Brin, Motwani, and Winograd}]{page:1999}
Lawrence Page, Sergey Brin, Rajeev Motwani, and Terry Winograd. 1999.
\newblock The pagerank citation ranking: Bringing order to the web.
\newblock Technical report, Stanford InfoLab.

\bibitem[{Peldszus and Stede(2015)}]{peldszus:2015}
Andreas Peldszus and Manfred Stede. 2015.
\newblock \href {https://doi.org/10.18653/v1/D15-1110} {Joint prediction in
  {MST}-style discourse parsing for argumentation mining}.
\newblock In \emph{Proceedings of the 2015 Conference on Empirical Methods in
  Natural Language Processing}, pages 938--948. Association for Computational
  Linguistics.

\bibitem[{Petasis and Karkaletsis(2016)}]{petasis:2016}
Georgios Petasis and Vangelis Karkaletsis. 2016.
\newblock \href {https://www.aclweb.org/anthology/W16-2811/} {Identifying
  argument components through textrank}.
\newblock In \emph{Proceedings of the Third Workshop on Argument Mining, hosted
  by the 54th Annual Meeting of the Association for Computational Linguistics,
  ArgMining@ACL 2016, August 12, Berlin, Germany}.

\bibitem[{Radford et~al.(2019)Radford, Wu, Child, Luan, Amodei, and
  Sutskever}]{radford:2019}
Alec Radford, Jeffrey Wu, Rewon Child, David Luan, Dario Amodei, and Ilya
  Sutskever. 2019.
\newblock Language models are unsupervised multitask learners.
\newblock \emph{OpenAI blog}, 1(8):9.

\bibitem[{Rothe et~al.(2020)Rothe, Narayan, and Severyn}]{rothe:2020}
Sascha Rothe, Shashi Narayan, and Aliaksei Severyn. 2020.
\newblock \href {https://transacl.org/ojs/index.php/tacl/article/view/1849}
  {Leveraging pre-trained checkpoints for sequence generation tasks}.
\newblock \emph{Trans. Assoc. Comput. Linguistics}, 8:264--280.

\bibitem[{Sanh et~al.(2019)Sanh, Debut, Chaumond, and Wolf}]{sanh:2019}
Victor Sanh, Lysandre Debut, Julien Chaumond, and Thomas Wolf. 2019.
\newblock Distilbert, a distilled version of bert: smaller, faster, cheaper and
  lighter.
\newblock \emph{arXiv preprint arXiv:1910.01108}.

\bibitem[{Schiller et~al.(2020)Schiller, Daxenberger, and
  Gurevych}]{schiller:2020}
Benjamin Schiller, Johannes Daxenberger, and Iryna Gurevych. 2020.
\newblock \href {https://arxiv.org/abs/2005.00084} {Aspect-controlled neural
  argument generation}.
\newblock \emph{CoRR}, abs/2005.00084.

\bibitem[{Shleifer and Rush(2020)}]{shleifer:2020}
Sam Shleifer and Alexander~M Rush. 2020.
\newblock Pre-trained summarization distillation.
\newblock \emph{arXiv preprint arXiv:2010.13002}.

\bibitem[{Stab and Gurevych(2014)}]{stab:2014}
Christian Stab and Iryna Gurevych. 2014.
\newblock \href {https://www.aclweb.org/anthology/D14-1006} {Identifying
  argumentative discourse structures in persuasive essays}.
\newblock In \emph{Proceedings of the 2014 Conference on Empirical Methods in
  Natural Language Processing ({EMNLP})}, pages 46--56, Doha, Qatar.
  Association for Computational Linguistics.

\bibitem[{Sutskever et~al.(2014)Sutskever, Vinyals, and Le}]{sutskever:2014}
Ilya Sutskever, Oriol Vinyals, and Quoc~V. Le. 2014.
\newblock \href
  {http://papers.nips.cc/paper/5346-sequence-to-sequence-learning-with-neural-networks}
  {Sequence to sequence learning with neural networks}.
\newblock In \emph{Advances in Neural Information Processing Systems 27: Annual
  Conference on Neural Information Processing Systems 2014, December 8-13 2014,
  Montreal, Quebec, Canada}, pages 3104--3112.

\bibitem[{Syed et~al.(2020)Syed, El~Baff, Kiesel, Al~Khatib, Stein, and
  Potthast}]{syed:2020}
Shahbaz Syed, Roxanne El~Baff, Johannes Kiesel, Khalid Al~Khatib, Benno Stein,
  and Martin Potthast. 2020.
\newblock \href {https://www.aclweb.org/anthology/2020.coling-main.470} {News
  editorials: Towards summarizing long argumentative texts}.
\newblock In \emph{Proceedings of the 28th International Conference on
  Computational Linguistics}, pages 5384--5396, Barcelona, Spain (Online).
  International Committee on Computational Linguistics.

\bibitem[{Toulmin(2003)}]{toulmin:2003}
Stephen~E Toulmin. 2003.
\newblock \emph{The uses of argument}.
\newblock Cambridge university press.

\bibitem[{Van~Dijk(1995)}]{dijk:1995}
Teun~A Van~Dijk. 1995.
\newblock Opinions and ideologies in editorials.
\newblock In \emph{4th International Symposium of Critical Discourse Analysis,
  Language, Social Life and Critical Thought, Athens}, pages 14--16.

\bibitem[{Wachsmuth et~al.(2017)Wachsmuth, Potthast, Al-Khatib, Ajjour,
  Puschmann, Qu, Dorsch, Morari, Bevendorff, and Stein}]{wachsmuth:2017}
Henning Wachsmuth, Martin Potthast, Khalid Al-Khatib, Yamen Ajjour, Jana
  Puschmann, Jiani Qu, Jonas Dorsch, Viorel Morari, Janek Bevendorff, and Benno
  Stein. 2017.
\newblock \href {https://doi.org/10.18653/v1/W17-5106} {Building an argument
  search engine for the web}.
\newblock In \emph{Proceedings of the 4th Workshop on Argument Mining}, pages
  49--59, Copenhagen, Denmark. Association for Computational Linguistics.

\bibitem[{Walton et~al.(2008)Walton, Reed, and Macagno}]{walton:2008}
Douglas Walton, Christopher Reed, and Fabrizio Macagno. 2008.
\newblock \emph{Argumentation Schemes}.
\newblock Cambridge University Press.

\bibitem[{Wang and Ling(2016)}]{wang:2016}
Lu~Wang and Wang Ling. 2016.
\newblock \href {https://doi.org/10.18653/v1/N16-1007} {Neural network-based
  abstract generation for opinions and arguments}.
\newblock In \emph{Proceedings of the 2016 Conference of the North {A}merican
  Chapter of the Association for Computational Linguistics: Human Language
  Technologies}, pages 47--57, San Diego, California. Association for
  Computational Linguistics.

\bibitem[{Wolf et~al.(2020)Wolf, Debut, Sanh, Chaumond, Delangue, Moi, Cistac,
  Rault, Louf, Funtowicz, Davison, Shleifer, von Platen, Ma, Jernite, Plu, Xu,
  Le~Scao, Gugger, Drame, Lhoest, and Rush}]{wolf:2020}
Thomas Wolf, Lysandre Debut, Victor Sanh, Julien Chaumond, Clement Delangue,
  Anthony Moi, Pierric Cistac, Tim Rault, Remi Louf, Morgan Funtowicz, Joe
  Davison, Sam Shleifer, Patrick von Platen, Clara Ma, Yacine Jernite, Julien
  Plu, Canwen Xu, Teven Le~Scao, Sylvain Gugger, Mariama Drame, Quentin Lhoest,
  and Alexander Rush. 2020.
\newblock \href {https://doi.org/10.18653/v1/2020.emnlp-demos.6} {Transformers:
  State-of-the-art natural language processing}.
\newblock In \emph{Proceedings of the 2020 Conference on Empirical Methods in
  Natural Language Processing: System Demonstrations}, pages 38--45, Online.
  Association for Computational Linguistics.

\bibitem[{Zhang et~al.(2019{\natexlab{a}})Zhang, Cai, Xu, and
  Wang}]{zhang:2019}
Haoyu Zhang, Jingjing Cai, Jianjun Xu, and Ji~Wang. 2019{\natexlab{a}}.
\newblock \href {https://doi.org/10.18653/v1/K19-1074} {Pretraining-based
  natural language generation for text summarization}.
\newblock In \emph{Proceedings of the 23rd Conference on Computational Natural
  Language Learning (CoNLL)}, pages 789--797, Hong Kong, China. Association for
  Computational Linguistics.

\bibitem[{Zhang et~al.(2020{\natexlab{a}})Zhang, Zhao, Saleh, and
  Liu}]{zhang:2020}
Jingqing Zhang, Yao Zhao, Mohammad Saleh, and Peter~J. Liu. 2020{\natexlab{a}}.
\newblock \href {http://proceedings.mlr.press/v119/zhang20ae.html} {{PEGASUS:}
  pre-training with extracted gap-sentences for abstractive summarization}.
\newblock In \emph{Proceedings of the 37th International Conference on Machine
  Learning, {ICML} 2020, 13-18 July 2020, Virtual Event}, volume 119 of
  \emph{Proceedings of Machine Learning Research}, pages 11328--11339. {PMLR}.

\bibitem[{Zhang et~al.(2020{\natexlab{b}})Zhang, Kishore, Wu, Weinberger, and
  Artzi}]{zhang:2020a}
Tianyi Zhang, Varsha Kishore, Felix Wu, Kilian~Q. Weinberger, and Yoav Artzi.
  2020{\natexlab{b}}.
\newblock \href {https://openreview.net/forum?id=SkeHuCVFDr} {Bertscore:
  Evaluating text generation with {BERT}}.
\newblock In \emph{8th International Conference on Learning Representations,
  {ICLR} 2020, Addis Ababa, Ethiopia, April 26-30, 2020}. OpenReview.net.

\bibitem[{Zhang et~al.(2019{\natexlab{b}})Zhang, Baldridge, and
  He}]{zhang:2019a}
Yuan Zhang, Jason Baldridge, and Luheng He. 2019{\natexlab{b}}.
\newblock \href {https://doi.org/10.18653/v1/N19-1131} {{PAWS}: Paraphrase
  adversaries from word scrambling}.
\newblock In \emph{Proceedings of the 2019 Conference of the North {A}merican
  Chapter of the Association for Computational Linguistics: Human Language
  Technologies, Volume 1 (Long and Short Papers)}, pages 1298--1308,
  Minneapolis, Minnesota. Association for Computational Linguistics.

\end{thebibliography}
